\documentclass[11pt]{article}

\usepackage[preprint]{acl}

\usepackage{times}
\usepackage{latexsym}

\usepackage[T1]{fontenc}
\usepackage[utf8]{inputenc}

\usepackage{microtype}

\usepackage{inconsolata}

\usepackage{graphicx}

\usepackage[english]{babel}
\usepackage{booktabs}
\usepackage{xcolor}
\usepackage{multirow}

\usepackage{tabularx}
\usepackage{array}

\usepackage{cuted}
\usepackage{caption}

\usepackage{amsmath}
\usepackage{amssymb}
\usepackage{float}
\usepackage{stfloats}
\usepackage[colorinlistoftodos]{todonotes}
\usepackage{natbib}
\title{Notation Matters: A Benchmark Study of Token-Optimized Formats in Agentic AI Systems}
\usepackage{authblk}

\author{Lorenz Kutschka \\
  Know Center Research GmbH / Sandgasse 34, A-8010 Graz \\
  \texttt{lkutschka@know-center.at} \\\And
  Bernhard C. Geiger \\
  Know Center Research GmbH / Sandgasse 34, A-8010 Graz \\
  \texttt{geiger@ieee.org}
}

\author{
 \textbf{Lorenz Kutschka\textsuperscript{1}},
 \textbf{Bernhard C. Geiger\textsuperscript{1,2,3}}
\\
 \textsuperscript{1}Know Center Research GmbH / Sandgasse 34, A-8010 Graz\\
 \textsuperscript{2}Signal Processing and Speech Communication, Graz University of Technology / Inffeldgasse 16c, A-8010 Graz\\
 \textsuperscript{3}Graz Center for Machine Learning // A-8010 Graz
\\
 \small{
   \textbf{Correspondence:} \href{mailto:lkutschka@know-center.at}{lkutschka@know-center.at}
 }
}

\begin{document}

\maketitle
\begin{abstract}
Large language models in Agentic AI systems consume tool schemas and execution results and emit tool invocations as structured data. The default language for that exchange, JSON, was designed for application-to-application interchange rather than token efficiency, so its structural elements impose substantial token overhead. Recent work proposes token-optimized alternatives such as TOON (Token-Oriented Object Notation) and TRON (Token Reduced Object Notation) as more compact replacements, but these formats have been evaluated only on isolated comprehension or generation tasks. Whether their token reductions hold inside end-to-end agentic loops therefore remains an open question. We evaluate TOON and TRON on four agentic benchmarks (BFCL, MCPToolBenchPP, MCP-Universe, StableToolBench) and five open-weight LLMs, decoupling input compression from output compression to measure comprehension and generation independently. TRON reduces tokens by up to 27\% with accuracy within 14~pp of the JSON baseline. TOON achieves up to 18\% reduction at a similar 9~pp accuracy cost, but additionally cascades on multi-turn parsing failures and collapses parallel tool-call output for most models. The code is available at: \href{https://github.com/lkutschka/notation-matters}{https://github.com/lkutschka/notation-matters}.
\end{abstract}

\begin{figure*}[t]
    \centering
    \includegraphics[width=0.8\textwidth]{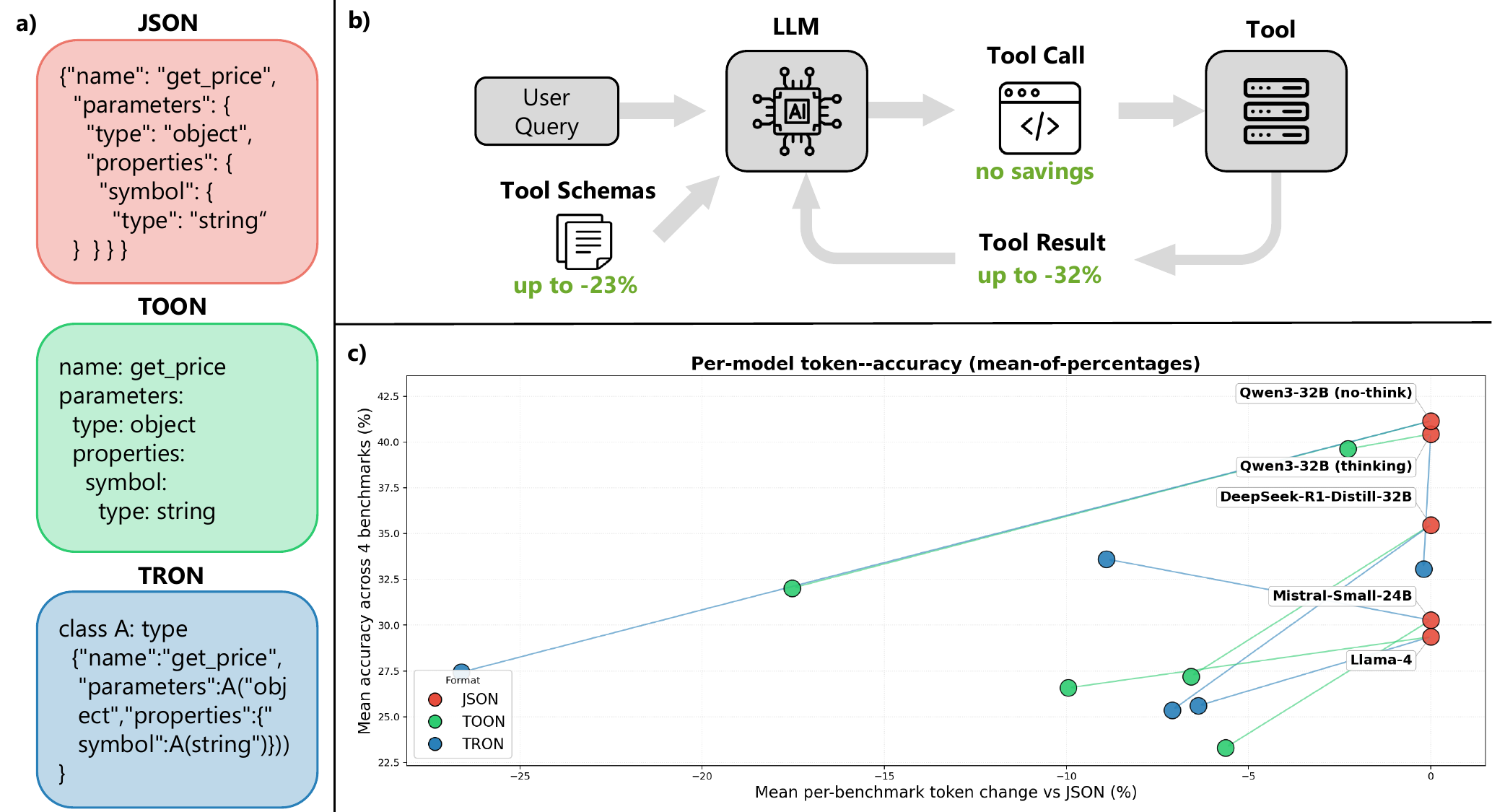}
    \caption{Format substitution in the tool-augmented pipeline. (a) The same tool schema encoded in JSON, TOON, and TRON. (b) The tool-calling loop, with green annotations marking the three format conversion points and the aggregate token saving observed at each. Tool schemas and tool results compress substantially (up to $-23\%$ and $-32\%$ respectively under TOON), while tool calls show no net saving due to parse failures in multi-turn benchmarks (see Sections~\ref{sec:results} and~\ref{sec:discussion}). (c) Per-model token--accuracy view, averaged across the four benchmarks (mean of per-benchmark percentages). Each marker is one (model, format) combination, with the x axis showing mean per-benchmark token change vs JSON and the y axis showing mean accuracy across benchmarks. Lines trace the JSON $\to$ TOON and JSON $\to$ TRON movement per model. TRON achieves up to $-27\%$ token reduction, further than TOON at $-18\%$, on Qwen3-32B no-think. Accuracy drops also vary widely: between $1$ and $9$~pp for TOON, and from a $+3$~pp gain to a $-14$~pp drop for TRON. Neither format dominates uniformly, and the format--model interaction is comparable in magnitude to the format choice itself.}

    \label{fig:pipeline_overview}
\end{figure*}

\section{Introduction}
\label{sec:introduction}
% Motivation
Large Language Models (LLMs) are increasingly deployed within Agentic AI systems, where they consume structured tool descriptions, interpret execution results, and emit tool invocations to accomplish user tasks. As these systems scale to more tools, longer descriptions, and richer result payloads, token consumption grows substantially. Because both inference cost and context-window limits are measured in tokens, the encoding of structured data is a direct determinant of system cost, energy efficiency, and feasibility.

% Research Gap
In current tool-augmented pipelines, JSON is the default data format for tool schemas, invocations, and results~\citep{Bourhis_Reutter_Suárez_Vrgoč_2017}, and LLMs interpret it effectively because of extensive exposure in their training corpora. JSON was, however, designed for application-to-application data interchange and not optimized for token efficiency, so its structural elements (quotation marks, commas, braces) impose substantial token overhead when consumed by LLMs. Recent work proposes token-optimized alternatives such as TOON (Token-Oriented Object Notation)~\citep{toon-format/toon_2026, Lafalce_2025} and TRON (Token Reduced Object Notation)~\citep{tron-format/tron-javascript_2026}, with initial results suggesting that LLMs can read TOON with minimal accuracy loss on isolated generation tasks~\citep{Masciari_Moscato_Napolitano_Orlando_Perillo_Russo_2026}. To date, however, no prior work has systematically evaluated these formats inside agentic pipelines that involve tool discovery, multi-turn reasoning, and execution.

% Objective
We address this gap by evaluating token-optimized alternatives to JSON (illustrated in Figure~\ref{fig:pipeline_overview}a) in agentic tool-augmented systems and measuring whether they reduce token consumption while preserving task accuracy.
% Method/Approach
Specifically, we benchmark TOON and TRON against JSON across function-calling and Model Context Protocol (MCP)-based benchmarks on five open-weight LLMs, substituting JSON at three points of the tool-calling loop (Figure~\ref{fig:pipeline_overview}b). We measure how much token reduction is achievable, how the substitution affects LLM accuracy in tool selection and parameter generation, and how both effects vary across model family, parameter scale, and architecture.

% Findings
Our results indicate that TRON reduces total tokens by up to 27\% while keeping accuracy within 14~pp of the JSON baseline across all four benchmarks. TOON achieves up to 18\% token reduction at a similar 9~pp accuracy cost, but its accuracy degrades further in multi-turn settings where parsing failures cascade into additional reasoning iterations and erode per-call gains. The token reductions scale with schema complexity, with the largest gains on benchmarks that expose many tools with detailed parameter definitions. Figure~\ref{fig:pipeline_overview}c shows the per-model token--accuracy tradeoff, and panel~b annotates where each reduction accrues across the three substitution points of the tool-calling loop.

% Impact
For deployments, our results identify TRON as a drop-in replacement for JSON in token-sensitive agentic systems, while TOON's stronger compression is offset by accuracy losses in multi-turn settings and is therefore not safe as a default. Researchers can use our evaluation framework to facilitate the evaluation of future serialization on the same comprehension and generation axes across four agentic benchmarks.
%Additional motivation paper: \citep{Levy_Jacoby_Goldberg_2024}
 
\section{Background}
Two background topics frame this study: the tool-calling loop through which LLMs interact with external systems, and the object-notation formats that encode the data flowing through it.

\subsection{Tool Augmentation}

Tool augmentation enables LLMs to query and modify external systems through a three-stage loop: the LLM receives structured tool definitions, generates a tool invocation, and receives the execution result. This loop runs once in single-turn tasks, where the model emits a single invocation and the interaction ends, and is iterated in multi-turn tasks, where the model interleaves reasoning with successive tool calls until producing a final answer. Across all implementations, JSON is the universal exchange format within Agentic AI. Current practice is shaped by two complementary standards: 1) function calling, the LLM-API convention for describing tools to the model and parsing its tool calls, and 2) the Model Context Protocol (MCP), the application-layer protocol that governs how applications discover and connect to remote tool servers.

\textbf{Function calling} is the convention in which tool definitions are provided as JSON Schema objects specifying a function name, description, and typed parameters. The LLM selects one or more functions and returns a structured invocation with the appropriate arguments. This convention originated with OpenAI's 2023 function-calling API and has since been adopted by most LLM providers and benchmarks. Evaluation typically relies on Abstract Syntax Tree (AST) matching, comparing the structure and values of the predicted invocation against a reference.

\textbf{MCP} is an open-source standard for connecting LLMs to external tools, databases, and systems, analogous to how HTTP standardized web communication~\citep{mcp2026}. MCP uses a client-server architecture in which servers host and execute tools or data resources. Clients are LLM-based applications that discover and trigger these tools to fulfill user requests. Communication is managed via a data layer using JSON-RPC for structured exchange and a transport layer that defines message delivery. Each server exposes its capabilities through an MCP schema consisting of three core primitives: tools (executable functions), resources (contextual information sources), and prompts (reusable templates). This standardization reduces integration overhead and provides predictable interfaces for tool discovery and invocation.

\subsection{Object Notation Formats}
\label{sec:object-notation-formats}

Tool schemas, invocations, and results are structured data, and the serialization format determines how that data appears as text in the prompt. We consider three formats. JSON is the universal baseline in current pipelines. TOON~\citep{toon-format/toon_2026} and TRON~\citep{tron-format/tron-javascript_2026} are two recently proposed formats that reduce token counts relative to JSON on the same data~\citep{Lafalce_2025, Masciari_Moscato_Napolitano_Orlando_Perillo_Russo_2026, Matveev_2026}. Figure~\ref{fig:serialization_examples} in Appendix~\ref{app:serialization_examples} shows the three formats applied to the same example data.

\paragraph{TOON.}

Token-Oriented Object Notation (TOON) is a serialization format designed as a lossless, drop-in alternative to JSON, optimized for token efficiency~\citep{toon-format/toon_2026}. TOON eliminates quotation marks, braces, and brackets by using YAML-like indentation for nested objects and CSV-style tabular notation for uniform arrays. This combination is particularly effective for structured data with repeated schemas, such as lists of objects sharing the same keys, and reduces token count substantially on such data~\citep{Lafalce_2025, Masciari_Moscato_Napolitano_Orlando_Perillo_Russo_2026, Matveev_2026}. The gains are largest on tabular structures and smaller for deeply nested or non-uniform data~\citep{Lafalce_2025}. However, structural-correctness rates degrade for models without native support~\citep{Masciari_Moscato_Napolitano_Orlando_Perillo_Russo_2026}, which is the caveat our agentic-loop evaluation builds on.

\paragraph{TRON.}

Token Reduced Object Notation (TRON) is a compact serialization format that reduces redundancy through class definitions~\citep{tron-format/tron-javascript_2026}. When multiple objects share the same structure, TRON defines the schema once as a named class and emits each object as a compact instance. Unlike TOON, TRON's compression scales with the number of repeated structures. This makes it effective for tool schemas, where multiple tools share similar parameter patterns. To exploit this scaling, tool schemas must be serialized as a single batch so that TRON can identify shared structures across tools. We refer to this requirement as the TRON batching condition. JSON and TOON do not benefit from such batching and are serialized individually.

\section{Related Work: Alternative Data Formats for LLMs}
\label{sec:related_work}

Several recent works have proposed token-optimized serialization formats as alternatives to JSON for structured data exchanged with LLMs. Lafalce~\citep{Lafalce_2025} provided a mathematical evaluation of TOON, showing that it reduces byte size compared to JSON, with particularly strong compression on tabular data. Masciari et al.~\citep{Masciari_Moscato_Napolitano_Orlando_Perillo_Russo_2026} extended this evaluation to generation tasks, benchmarking TOON against JSON, XML, and YAML across multiple LLMs. Their results show that TOON yields more compact outputs and lower carbon emissions, but structural correctness decreases for models that were not exposed to TOON during pre-training, with larger models closing this gap. 

The accuracy impact of TOON is highly model-dependent. McMillan~\citep{McMillan_2026} ran 9,649 prompt-completion trials across 11 models and 4 formats (YAML, Markdown, JSON, TOON) on SQL generation tasks, finding that format does not significantly affect aggregate accuracy ($\chi^2$=2.45, p=0.484), though individual models show format-specific sensitivities with accuracy changes ranging from \mbox{-7.7\%} to \mbox{+2.7\%}. Consistent with Masciari et al.~\citep{Masciari_Moscato_Napolitano_Orlando_Perillo_Russo_2026}, the dominant factor was model capability, with a 21~pp gap between frontier and open-source tiers, dwarfing any format effect. Matveev~\citep{Matveev_2026} compared TOON generation against plain JSON and constrained-decoding JSON, finding that TOON has a favorable accuracy-to-token ratio for in-domain tasks but that the prompt overhead needed to teach the model TOON syntax can wipe out the per-token savings on short outputs. TOON therefore only becomes cost-effective once the output is large enough for cumulative per-token savings to outweigh the fixed prompt cost.

Concurrent work has explored further JSON alternatives. JTON~\citep{Nandakishore_2026}, like TRON, factors shared structure across repeated objects through a Zen Grid tabular encoding. It reports 15--60\% token reduction across seven domains, with comprehension and one-shot generation tests on 10--12 LLMs. ONTO~\citep{Deekeswar_2026} uses pipe-delimited columnar notation declaring each entity's fields once, reducing tokens by 46--51\% on IoT data, with comprehension verified on Qwen2.5-7B. Neither JTON nor ONTO are evaluated inside multi-turn agentic loops, and JTON's batching is restricted to homogeneous arrays whereas TRON generalizes to repeated structural patterns across heterogeneous tool definitions.

Beyond general-purpose formats, domain-specific notations have emerged for time-series data. TSLN~\citep{Mudbari_Bhagat_2026} achieves 68--73\% token reduction through schema-first architecture, relative timestamps, and differential encoding. TOKON~\citep{Yang_2025} reduces tokens by 2--3$\times$ through tokenization-optimized normalization, improving forecasting root mean square error by 7--18\%. Both TSLN and TOKON are orthogonal to our work: they target time-series payloads specifically, whereas tool-augmented pipelines exchange heterogeneous tool-schema, tool-call, and tool-result data. Alshaer~\citep{Alshaer} identified that TOON's lack of explicit delimiters allows attacker-controlled string content to be reparsed as schema fields, and proposed the S-TOON protocol, which reintroduces explicit boundary markers around untrusted input. Although our work does not target adversarial inputs, the existence of this attack underlines that format choice has consequences beyond efficiency.

The works above show that token-optimized formats can reduce token consumption with varying accuracy trade-offs. Several of them already evaluate format substitution systematically across multiple LLMs and formats~\citep{Masciari_Moscato_Napolitano_Orlando_Perillo_Russo_2026, McMillan_2026, Matveev_2026}, but only for isolated comprehension or generation tasks. 

Our work is, to our knowledge, the first systematic evaluation of token-optimized formats inside agentic tool-calling pipelines, the first to decouple input-side compression from output-side compression as separate experimental conditions, and the first to report TRON's effect on LLM accuracy.

\section{Experimental Setup}
\label{sec:experimental_setup}

To evaluate whether LLMs can both consume and produce token-optimized formats inside agentic pipelines, we substitute JSON with TOON and TRON at three points of the tool-calling loop:

% \begin{enumerate}
% \vspace{5pt}
    \textbf{1) Tool schemas (input):} The list of available tool definitions, including each tool's name, description, and typed parameter specification, is serialized in the target format and included in the system prompt at session start. This block is typically the largest single piece of structured data in the prompt, since it scales with the number of available tools.
    
% \vspace{5pt}
    \textbf{2) Tool calls (output):} At each agent step, the LLM emits a structured invocation in the target format (JSON, TOON, TRON) specifying the tools to call and its arguments. The agent deserializes this and re-encodes the arguments as JSON before invoking the tool, since function-calling APIs and MCP's JSON-RPC transport both require JSON-encoded arguments.
    
% \vspace{5pt}
    \textbf{3) Tool results (input):} Execution results returned by tools or MCP servers are serialized in the target format and appended to the conversation context for the next reasoning step. Results grow across iterations in multi-turn execution, since each tool call appends fresh content to the conversation context for the next reasoning step.
% \end{enumerate}

\vspace{5pt}
% \paragraph{Experimental Conditions.}
From the LLM's perspective, we decouple input and output formats. The input format governs how tool schemas and results are presented to the LLM, while the output format governs how the LLM must structure its tool invocations. This decoupling enables two experimental conditions. First, \textbf{input-only} compression, where the LLM consumes schemas and results in the target format but still emits tool calls as JSON. Second, \textbf{full} compression, where the LLM works in the target format end-to-end and also emits tool calls in the same format. All format conversions are routed through a unified serialization library (shared\_format) that centralizes encoding and decoding behavior across the four benchmarks. At every substitution point, the LLM emits the target format as a stand-alone document: the entire per-turn response is a single TOON or TRON object with explicit \texttt{thought}, \texttt{action}, and \texttt{arguments} keys, rather than wrapped inside a labeled plain-text field such as the ReAct convention's \texttt{Action Input}~\citep{yao2023react} (the Action-Input anti-pattern, Appendix~\ref{app:benchmark_details}).

\subsection{Benchmarks}
\label{sec:benchmarks}

We use four tool-augmented benchmarks that together cover  function calling and MCP, each in a single-turn and a multi-turn setting. BFCL~\citep{patil2023gorilla} and StableToolBench~\citep{Guo_Cheng_Wang_Liang_Qin_Li_Liu_Sun_Liu_2025} provide single- and multi-turn function calling, and MCPToolBenchPP~\citep{fan2025mcptoolbench} and MCP-Universe~\citep{luo2025mcpuniverse} provide the same two settings for MCP. We run all four benchmarks in prompt mode, since the native function-calling APIs of inference providers serialize tool schemas internally and emit tool calls only as JSON, which prevents the format substitution we study.

These four benchmarks offer diversity in multiple characteristics, that are essential to our evaluation: the shape of data, the structure of the interaction, tool-catalog size, result payload depth and single-/multi-turn settings. Appendix~\ref{app:benchmark_comparison} compares these benchmarks and explains their diversity in the mentioned aspects.

\subsection{Evaluation Metrics}
\label{sec:metrics}

\paragraph{Token Consumption Metrics.}
We decompose token consumption into schema tokens, tool-call tokens, and result tokens, all measured with the cl100k\_base tokenizer from tiktoken. Total prompt and completion tokens per call are reported by the inference API (vLLM for self-hosted runs). Token-consumption metrics are defined consistently across all four benchmarks. For cross-benchmark comparison we also report each format's relative change against the JSON baseline as a percentage gain or loss, which yields a benchmark-agnostic measure of format effect.

\paragraph{Accuracy Metrics.}
Task accuracy uses each benchmark's own standard evaluation, since each benchmark defines a task-appropriate notion of correctness. BFCL uses AST-matching accuracy against a reference invocation. MCPToolBenchPP reports a hierarchical pass@1 over five trials, with overall pass@1, tool\_pass@1 for tool selection, and param\_pass@1 for parameter correctness given a correct tool. MCP-Universe reports a strict pass rate, where a task counts as solved only if all of its evaluators succeed. StableToolBench reports the Solvable Pass Rate (SoPR), the fraction of solvable queries judged to fulfill the user's intent.

\subsection{Model Configuration}
\label{sec:model}

Format sensitivity may depend on the model. To check whether observed effects are general or model-specific, we evaluate five open-weight LLM configurations against all four benchmarks. The five configurations span four model families (Qwen3, DeepSeek, Mistral, Llama) and vary along four axes: parameter scale (17B--32B active parameters), architecture (dense versus Mixture-of-Experts), training emphasis (Qwen3 and Mistral are explicitly trained for tool calling, while DeepSeek-R1-Distill is reasoning-distilled from Qwen-2.5-32B, and Llama-4-Scout is the MoE variant in the Llama-4 family), and reasoning mode (Qwen3-32B is evaluated with thinking on and off as two separate configurations). Table~\ref{tab:models} in Appendix~\ref{app:models} lists the full configuration.

\section{Results}
\label{sec:results}

We address the three research questions across the format--model matrix:
\begin{itemize}
    \item[RQ1] How much token reduction is achievable when replacing JSON with token-optimized alternatives in agentic tool-augmented systems?
    \item[RQ2] Does format substitution affect LLM accuracy in tool selection and parameter generation?
    \item[RQ3] How does the effect of token-optimized formats vary across open-weight LLMs that differ in family, parameter scale, and architecture?
\end{itemize}

\begin{figure}[!htbp]
    \centering
    \includegraphics[width=\columnwidth]{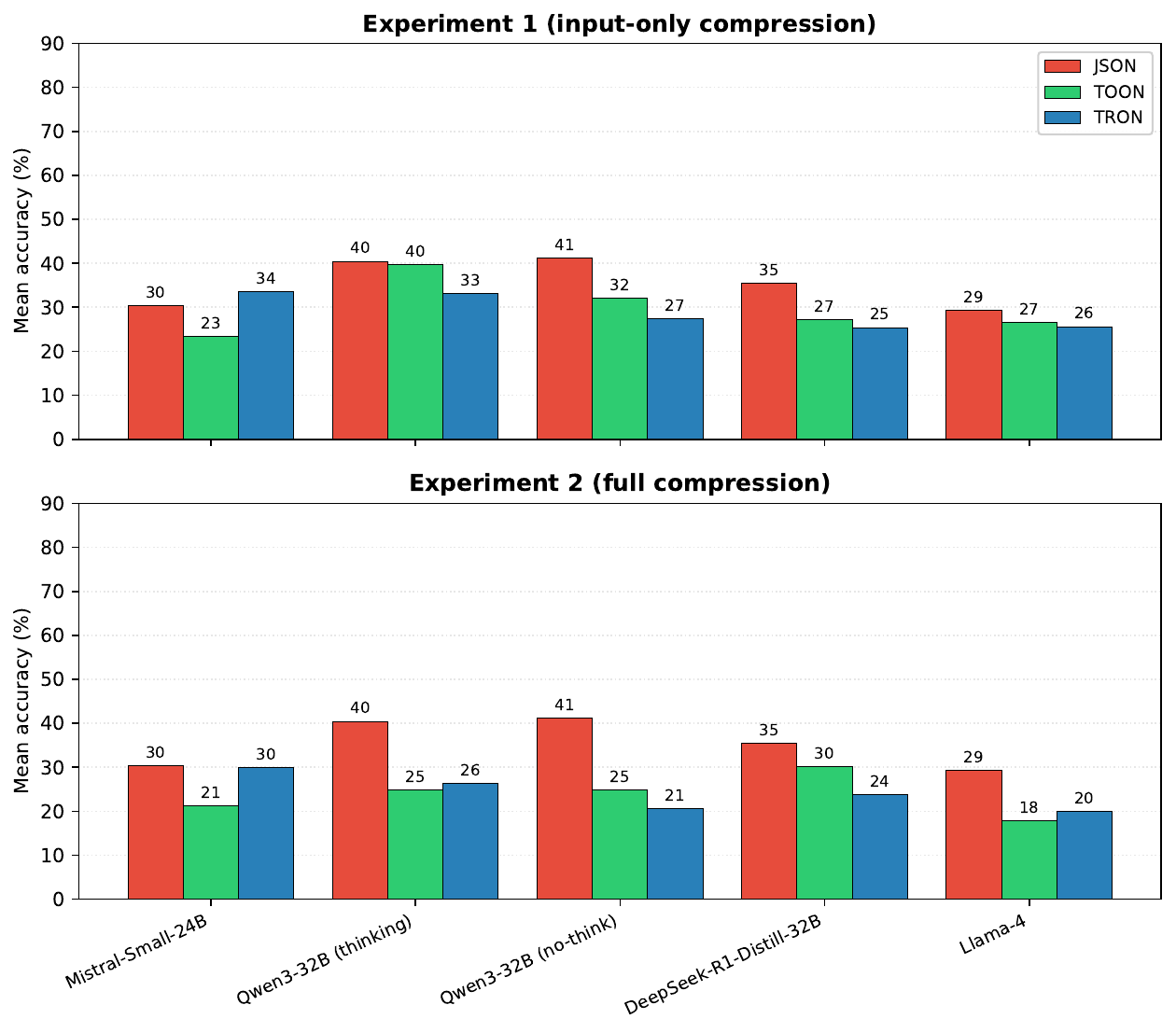}
    \caption{Mean accuracy per model and format, averaged across benchmarks. Each triple of grouped bars compares JSON, TOON, and TRON within one model. Top: \textbf{input-only} compression. Bottom: \textbf{full} compression, including tool calls. In general, all models suffer accuracy losses under compression, with TOON dropping more than TRON. Exceptions are Qwen3-32B with thinking whose TOON accuracy matches its JSON baseline under input-only compression, and Mistral-Small-24B, which achieves JSON accuracy with TRON.}
    \label{fig:accuracy_by_model}
\end{figure}

\begin{figure*}[!htbp]
    \centering
    \includegraphics[width=0.95\textwidth]{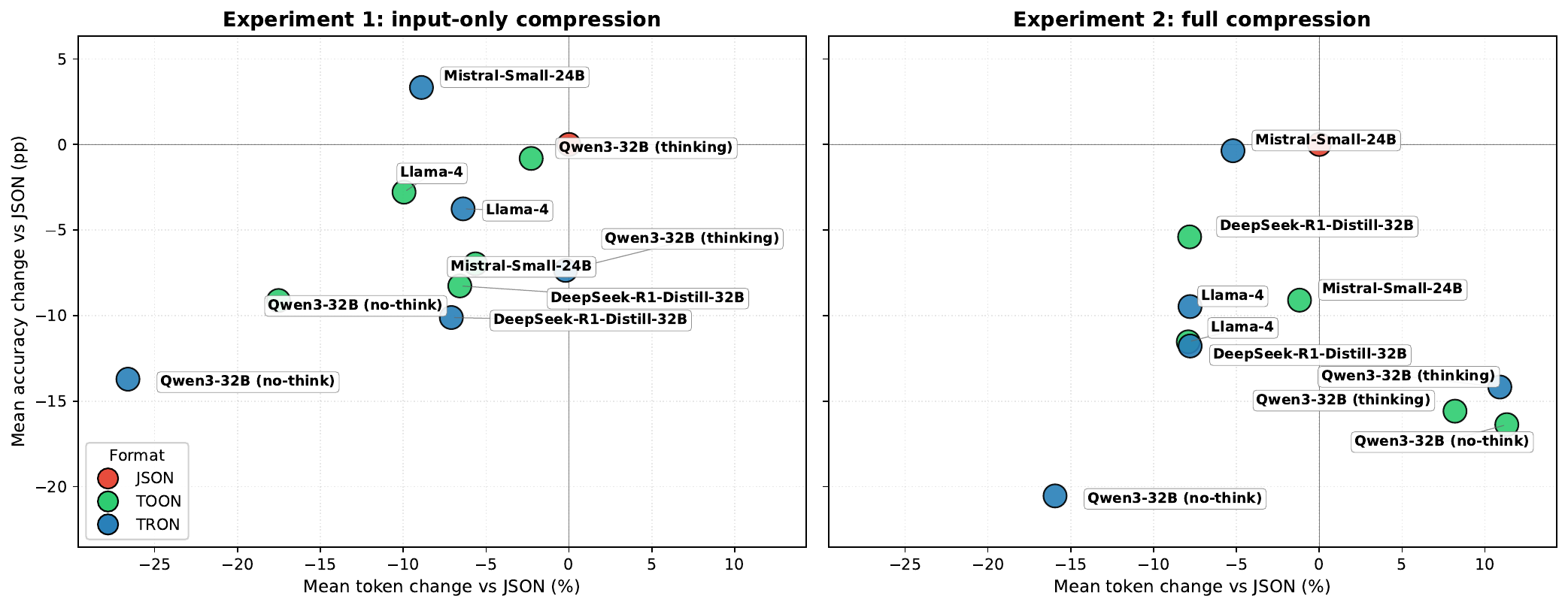}
    \caption{Per-model token--accuracy tradeoff against the JSON baseline, averaged across benchmarks. \textbf{Left:} input-only compression. \textbf{Right:} full compression, including tool calls. Each marker is one (model, format) combination, and the red JSON marker sits at the origin in both panels. Under input-only compression, format effects are model-dependent: Mistral-Small-24B gains accuracy under TRON, while reasoning configurations (Qwen3-32B no-think, DeepSeek-R1-Distill) lose more accuracy on TRON than on TOON. Under full compression, several configurations drift to the right of the origin because the parsing-cascade pushes total tokens above the JSON baseline, and accuracy losses widen for both formats.}
    \label{fig:tradeoff_scatter_both}
\end{figure*}

We evaluate two settings. \textbf{Experiment 1 (input-only compression)} substitutes JSON with TOON or TRON for schemas and results while tool calls remain JSON, isolating format comprehension from generation. \textbf{Experiment 2 (full compression)} additionally emits tool calls in the target format, testing end-to-end production. Figure~\ref{fig:accuracy_by_model} reports mean accuracy per model and format for both settings, and Figure~\ref{fig:tradeoff_scatter_both} reports the per-model token--accuracy tradeoff. Per-cell numbers are in Tables~\ref{tab:accuracy_input_only} and~\ref{tab:accuracy_full}.

\paragraph{Token reduction (RQ1).} Under Experiment 1, averaged per model across the four benchmarks, TOON reduced total tokens by 2--18\% and TRON by 0--27\% (Figure~\ref{fig:tradeoff_scatter_both}, left), with both maxima reached on Qwen3-32B (no-think). The TRON batching condition (Section~\ref{sec:object-notation-formats}) determines where TRON's savings concentrate: the class-definition mechanism delivers consistent reductions on benchmarks whose tasks expose many tools per query (MCPToolBenchPP saves roughly 10\% per model, StableToolBench up to 22\%) but \emph{increases} tokens by up to 21\% on BFCL's non-live categories, where each task references only a few unique tool schemas and the per-batch class header costs more than the body it deduplicates. Under Experiment 2, adding output-side compression does not uniformly increase savings. Non-reasoning configurations (Mistral-Small-24B, Llama-4-Scout, DeepSeek-R1-Distill) continued to save 1--10\% of total tokens on average, but the reasoning configurations (Qwen3-32B with and without thinking) drifted toward or above the JSON baseline (Figure~\ref{fig:tradeoff_scatter_both}, right). The drift is concentrated on the multi-turn benchmarks (MCP-Universe and StableToolBench), where the parsing-cascade applies: a failed parse of a tool call triggers an additional reasoning iteration whose own thought, action, and observation payload outweighs the per-call savings. After partial offset by savings on the single-turn benchmarks, the per-model means resolve to $+8\%$ (Qwen3-32B with thinking, TOON), $+11\%$ (Qwen3-32B with thinking, TRON), and $+11\%$ (Qwen3-32B without thinking, TOON).

\paragraph{Accuracy (RQ2).} Under Experiment 1, TRON's accuracy averaged across the four benchmarks varied between $+3$~pp on Mistral-Small-24B and $-14$~pp on Qwen3-32B (no-think) (Figure~\ref{fig:tradeoff_scatter_both}). TOON's accuracy varied more widely. The single largest drop was Mistral-Small-24B on BFCL's non-live categories, where AST-matching accuracy fell from 89\% under JSON to 53\% under TOON (visible as the lower-left Mistral point in the BFCL panel of Figure~\ref{fig:faceted_scatter_all}, Table~\ref{tab:accuracy_input_only}). We hypothesise that smaller models with limited pre-training exposure to TOON syntax are particularly vulnerable on BFCL's strict AST evaluation, consistent with the structural-correctness gap that \citet{Masciari_Moscato_Napolitano_Orlando_Perillo_Russo_2026} report for smaller models without native TOON support. The same Mistral configuration, however, \emph{gains} accuracy on MCPToolBenchPP under both TOON ($+13$~pp at $-29\%$ tokens) and TRON ($+17$~pp at $-10\%$ tokens), suggesting that the format's effect interacts strongly with task type rather than acting uniformly. Under Experiment 2, full compression exposes a sharp degradation on BFCL's \texttt{parallel} and \texttt{parallel\_multiple} categories (\texttt{parallel}: the same function called multiple times in parallel, \texttt{parallel\_multiple}: several different functions called in parallel), where accuracy collapses to near zero for most (model, format) combinations. This is the most significant negative finding of the study: our models, which are pre-trained predominantly on JSON tool-call output, do not reliably generate parallel invocations in alternative formats because the structural cues that disambiguate sequential from parallel calls are not consistently preserved when the format is substituted. On the multi-turn benchmarks the parsing-cascade compounds under full compression, because parsing failures during the output stage themselves trigger further reasoning iterations. DeepSeek-R1-Distill-32B is the only configuration whose mean TOON accuracy improved under Experiment 2 relative to Experiment 1 ($+3$~pp), driven entirely by StableToolBench where its SoPR jumped from 22\% to 46\%. We flag this as an outlier without a mechanistic explanation.

\paragraph{Per-model variation (RQ3).} Under Experiment 1, Qwen3-32B with thinking enabled was the only configuration whose TOON accuracy matched its JSON baseline. Toggling thinking off on the same base model dropped TOON accuracy by 8~pp and TRON accuracy by 6~pp, indicating that the explicit reasoning step helps the model recover from format-driven structural surprises and that model size and family are not the only drivers for robustness. The other reasoning-capable configuration, DeepSeek-R1-Distill-32B, did not inherit this robustness: its TRON accuracy dropped by 10~pp on average, the second-largest TRON drop in the matrix after Qwen3-32B (no-think, Table~\ref{tab:accuracy_input_only}). Llama-4-Scout-17B's MoE architecture did not show a systematic advantage on input compression over comparable dense models. Under Experiment 2, the thinking-mode advantage changes character. For TRON output, the gap between Qwen3-32B with and without thinking persists at 5~pp (26\% versus 21\% mean accuracy), but for TOON output the gap vanishes entirely: both configurations land at 25\%. Explicit reasoning thus helps the model generate well-formed TRON under full compression but does not rescue its TOON generation. Llama-4-Scout's MoE architecture again showed no systematic advantage on alternative output formats. Across the five configurations, no single model succeeded under full compression: every configuration lost at least 4~pp of accuracy under TOON, and only Mistral-Small-24B retained TRON accuracy within 0.4~pp of JSON.

\section{Discussion}
\label{sec:discussion}

\paragraph{Token cost: when compression delivers and when it backfires (RQ1).}
The token effects of TOON and TRON in our matrix are governed by two interacting mechanisms. TRON's class-definition mechanism only pays off when the serialization batch contains repeated structural patterns. In our setup, this condition is met when a task exposes many tools whose parameter schemas share fields. This is the typical shape of MCP servers and of StableToolBench's API bundles, and TRON reduces tokens consistently on these benchmarks (up to 27\% on Qwen3-32B no-think). The same mechanism becomes a liability on BFCL's non-live categories. Each task there references only a handful of unique tool schemas, and the per-batch class header costs more than the body it deduplicates, so total tokens \emph{increase} by up to 21\%. A second backfire mechanism is specific to multi-turn pipelines: a parse failure on a single tool call triggers an additional reasoning iteration, and that iteration carries its own thought, action, and observation payload. For reasoning configurations this extra payload includes the model's full chain of thought. Two or three triggered iterations are then sufficient to consume the per-call savings accumulated over an entire trajectory. Qwen3-32B with thinking ends at $+8\%$ tokens under full TOON and $+11\%$ under full TRON, and Qwen3-32B without thinking at $+11\%$ under TOON (Tables~\ref{tab:accuracy_input_only} and~\ref{tab:accuracy_full}). Whether a token-optimized format saves tokens therefore depends on the shape of the workload and on the model's parse-success rate, not on per-call format compression in isolation.

\paragraph{Accuracy can improve under compression (RQ2).}
Several configurations in our matrix improved accuracy under compression rather than degrading. The clearest case is Mistral-Small-24B on MCPToolBenchPP, where accuracy rose by $+13$~pp under TOON at $-29\%$ tokens and by $+17$~pp under TRON at $-10\%$ tokens (visible in the MCPToolBenchPP panel of Figure~\ref{fig:faceted_scatter_all}; per-cell numbers in Table~\ref{tab:accuracy_input_only}). We hypothesise that compression reduces noise on the model's attention surface, freeing attention budget for the task content. Removing JSON's structural elements (braces, brackets, repeated key strings) drops a substantial share of tokens that carry no task-relevant signal. This reading is consistent with the long-context literature, which reports smaller models benefiting from denser prompts more than larger ones. It also explains why the gain was largest on the smallest model in our matrix and on the benchmark with the most verbose JSON schemas. The effect was not uniform across the matrix, so we treat it as a hypothesis for follow-up rather than as a deployment recommendation.

\paragraph{Accuracy losses follow a coherent pattern (RQ2).}
The accuracy losses we observe under compression cluster on two failure modes. Under input-only compression, the losses concentrate on models whose pre-training exposure to the target format is presumably low, with the largest single drop on Mistral-Small-24B's BFCL accuracy (89\% under JSON to 53\% under TOON, Table~\ref{tab:accuracy_input_only}). Under full compression the failure mode shifts from comprehension to generation. BFCL's \texttt{parallel} and \texttt{parallel\_multiple} categories collapse to near-zero accuracy across most (model, format) combinations, suggesting that LLMs trained predominantly on JSON tool-call output do not reliably reproduce the parallel-call structure in alternative formats. The Action-Input anti-pattern surfaced during pilot runs reinforces this reading: when TOON content was embedded inside a labelled plain-text field, models defaulted back to JSON-shaped output regardless of the surrounding format instruction. The shared mechanism is that JSON output is the deeper prior, and alternative formats require either explicit reasoning at inference time or explicit training-time exposure.

\paragraph{Training emphasis and reasoning mode (RQ3).}
The five model configurations differ along four axes (parameter scale, architecture, training emphasis, reasoning mode), and the dominant variable in our matrix is reasoning mode. Qwen3-32B with thinking enabled is the only configuration whose TOON accuracy matches its JSON baseline under input-only compression. Toggling thinking off on the same base model drops TOON accuracy by 8~pp and TRON accuracy by 6~pp. We hypothesise that explicit reasoning lets the model recover when it encounters an unfamiliar input format, by reasoning over the structure before committing to an answer. The other reasoning-capable configuration, DeepSeek-R1-Distill-32B, did not inherit this robustness and dropped 10~pp on TRON (Table~\ref{tab:accuracy_input_only}). This suggests that distillation from Qwen-2.5 does not transfer format-side robustness, and that explicit chain-of-thought at inference time is the load-bearing variable. Llama-4-Scout-17B's MoE architecture did not show a systematic advantage over comparable dense models, indicating that MoE routing did not select format-specialised experts on our matrix. Tool-calling training emphasis, present in both Qwen3 and Mistral, did not protect Mistral from BFCL's strict AST evaluation under TOON, indicating that tool-call training transfers to comprehension but not to robust generation in unfamiliar formats. The pattern across the four axes is that no single architectural or training choice is decisive on its own, and that an inference-time reasoning step is the only variable that yields a consistent advantage in our matrix.

\paragraph{Practical guidance.}
For deployments, our results identify TRON as a defensible drop-in for JSON in token-sensitive agentic systems when the workload exposes many structurally similar tools, and otherwise as a workload-dependent choice that should be measured rather than assumed. TOON is not safe as a default in multi-turn agentic systems, because its stronger per-call compression is offset by parsing-cascade effects and by structural failures on parallel tool-call output.

\section{Conclusions}

% Contribution
We evaluated TOON and TRON inside agentic tool-calling pipelines on four benchmarks and five open-weight LLM configurations, decoupling the input and output sides of the loop so that comprehension and generation could be measured independently. TRON behaves as a drop-in replacement for JSON under input-only compression, reducing total tokens by up to 27\% with accuracy within 14~pp of the JSON baseline, provided the workload exposes enough structurally similar tools for its class-definition mechanism to amortize. TOON yields up to 18\% reduction at a similar 9~pp accuracy cost. However, TOON loses accuracy further in multi-turn settings, where parsing failures cascade into additional reasoning iterations and erode per-call gains. Follow-up work should characterise the interaction between format substitution and tool-call training emphasis on frontier-scale models, and evaluate whether targeted post-training on tool-call outputs in alternative formats removes the structural failures we observe under full compression.

\clearpage

\section*{Limitations}

\paragraph{Model scope.} We evaluated five open-weight LLM configurations spanning 17B--32B active parameters and four model families (Qwen3, DeepSeek, Mistral, Llama). We restricted the evaluation to open-weight models to keep weights and pre-training data inspectable, which makes contamination analysis and replication tractable. This excluded the largest closed-source frontier models (e.g., GPT-5, Claude Opus 4.7, Gemini 3.1 Pro), which are not self-hostable and whose API costs for the full format-model-experiment matrix would have exceeded our research budget. Among open-weight models, large dense models above 100B parameters exceeded the GPU memory available to us for this study. Format sensitivity at frontier scale may therefore differ from what we reported. In particular, larger models may close the structural-correctness gap that smaller models exhibit on TOON output~\citep{Masciari_Moscato_Napolitano_Orlando_Perillo_Russo_2026}.

\paragraph{Combined schema and result compression.} Our setup for input-only compression compressed both tool schemas (input at session start) and tool results (input per turn) in the target format. We did not separately measure schema-only versus result-only compression. The two inputs have different shapes: schemas are large and structurally repetitive, while results are smaller but vary widely in nesting depth, so the format effects may differ between them. A finer-grained ablation isolating the contributions of schema-side and result-side compression is left to future work.

\paragraph{Excluded benchmark categories.} Two benchmarks contain categories that we excluded because of external-service constraints. MCP-Universe's \texttt{web\_search} category required a SERP API budget beyond what was available for repeated runs across the format and model matrix. MCPToolBenchPP's \texttt{search} and \texttt{browser} categories depended on third-party APIs (Tavily, Playwright) that returned quota errors or launch failures uniformly across formats. We therefore reported 5 of 6 MCP-Universe categories and 4 of 6 MCPToolBenchPP categories. Appendix~\ref{app:benchmark_details} documents the per-benchmark scope.

\paragraph{Measurement caveats.} On MCPToolBenchPP, the LLM-judge rejudge stage produced JSON-format parse failures on approximately 3.6\% of trials (measured on the Mistral-Small-24B baseline). This floor applies uniformly across formats and therefore biases all MCPToolBenchPP point estimates downward by a similar margin, without affecting relative comparisons between formats. Other, model-specific issues were resolved at the parsing layer and are documented in Appendix~\ref{app:implementation_notes}.

\section*{Ethics Statement}
\label{sec:ethics}

\paragraph{Compute cost.} Our work motivates inference-time token savings, and tokens map to compute and to the energy footprint of inference. The evaluation itself required substantial compute across four benchmarks, five open-weight LLM configurations, three formats, and two experimental conditions. We acknowledge the tension between studying token-side efficiency and the compute spent on the study, and view the per-deployment savings projected by our findings as the appropriate frame for amortizing this initial cost.

\paragraph{Open-weight models.} Our evaluation is restricted to open-weight LLMs (Section~\ref{sec:model}) both for reproducibility and for the auditability afforded by inspectable weights. Closed-source APIs can change their underlying models, system prompts, and moderation policies without notice, and their pre-training data is not disclosed, making both replication and contamination analysis impossible. Open-weight models do not eliminate these concerns but make them tractable to study.

\paragraph{Out-of-scope ethical dimensions.} The benchmarks we use evaluate tool-calling correctness only. They do not measure fairness across user demographics, bias in tool selection, or harm from misaligned tool use. These dimensions are outside the scope of this study, and practitioners considering deployment of token-compressed pipelines should evaluate them separately on top of our methodology.

\section{Acknowledgements}
This work is funded under the Austrian COMET Programme (COMET Module IACAI - INTERFACES OF AGENT-CENTRIC AI), supported by the Federal Ministry of Innovation, Mobility and Infrastructure (BMIMI), Federal Ministry of Economy, Energy and Tourism (BMWET), the federal province of Styria, and managed by the FFG. The computational results have been achieved using the Austrian Scientific Computing (ASC) infrastructure. We thank Hussain Hussain for insightful discussions and valuable feedback on this work.

\bibliography{bibliography}

\clearpage

\appendix
\section*{Appendix}
\addcontentsline{toc}{section}{Appendix}

\section{Benchmark Comparison}
\label{app:benchmark_comparison}
Table~\ref{tab:mcp_benchmarks} compares tool-augmented benchmarks considered for this study, and highlights those that were selected.

The four-way coverage in our study is necessary because format efficiency depends on both the shape of the data and the structure of the interaction. TOON achieves its gains on uniform arrays, while TRON exploits repeated tool-parameter shapes through class-based instantiation. Either property may be absent from a given task suite, so a single benchmark would risk reporting savings that do not generalize. The four benchmarks also differ in tool-catalog size and in result-payload depth, exercising compression across heterogeneous data structures rather than a single representative one. Multi-turn settings also expose a parsing-cascade effect that single-turn measurements miss: parsing failures trigger further reasoning iterations and erode per-call gains. Including both function calling and MCP further isolates the effect of the encoding itself from any artifact of one ecosystem's schema conventions.

\begin{table*}[h]
    \centering
    \footnotesize
    \setlength{\tabcolsep}{4pt}
    \renewcommand{\arraystretch}{1.15}
    \caption{Comparison of tool-augmented benchmarks considered for this study. \textbf{Eval Method} distinguishes between subjective LLM judges and objective execution verification. Bold rows indicate benchmarks selected for this study.}
    \label{tab:mcp_benchmarks}

    \begin{tabularx}{\textwidth}{@{}l>{\raggedright\arraybackslash}X>{\raggedright\arraybackslash}X>{\raggedright\arraybackslash}X@{}}
        \toprule
        \textbf{Benchmark} & \textbf{Scale \& Domain} & \textbf{Eval Method} & \textbf{Process Metrics} \\
        \midrule

        \textbf{MCPToolBenchPP} &
        1,509 tasks; 6 domains (finance, maps, browser) &
        \textbf{Hybrid} (AST match + LLM judge) &
        Pass@k; token decomposition \\

        \addlinespace
        \textbf{MCP-Universe} &
        231 tasks; 6 domains (finance, web, 3D) &
        \textbf{Execution} (live API state verification) &
        Tokens/turn; iteration count \\

        \addlinespace
        \textbf{BFCL v4} &
        4,447 tasks; synthetic + live APIs, 17 categories &
        \textbf{AST match} (deterministic) &
        Overall Acc; per-category \\

        \addlinespace
        \textbf{StableToolBench} &
        330 solvable queries; 7,000+ cached real APIs &
        \textbf{LLM judge} (GPT-4, SoPR + SoWR) &
        Pass rate; iterations; tokens \\

        \addlinespace
        MCPMark &
        127 tasks; 5 servers (CRUD, DB) &
        Execution (state check) &
        Pass$^4$ (retries); cost \\

        \addlinespace
        MCPAgentBench &
        841 tasks; 20k+ tools (simulated) &
        Exact match (golden tool set) &
        Token efficiency (score/1k tok) \\

        \addlinespace
        MCP-Bench &
        Synthesized; 28 servers (scientific) &
        Hybrid (LLM judge + schema) &
        Planning score (qualitative) \\

        \bottomrule
    \end{tabularx}
\end{table*}

\section{Benchmark Details}
\label{app:benchmark_details}

A common design choice runs across all four benchmarks. None of them use the native function-calling APIs offered by inference providers, because those APIs handle tool-schema serialization internally and accept tool calls only as JSON. To substitute TOON or TRON we therefore embed schemas and (where applicable) result observations as text in the system prompt. Each benchmark's prompt also contains a worked example showing the syntactic shape of a valid response in the target format, and the LLM is instructed to emit its response (or the format-specific portion of it, in the case of MCPToolBenchPP) entirely in that format rather than only a specific field. This avoids what we refer to as the Action-Input anti-pattern, in which a format block is embedded inside a labeled plain-text field. Our pilot runs found that this anti-pattern triggers the model's JSON-tool-call prior and collapses adoption of the target format.

\subsection{MCPToolBenchPP}
MCPToolBenchPP \citep{fan2025mcptoolbench} evaluates single-turn tool calling. Given a natural language query and a set of tool schemas, the LLM must produce a single tool invocation with the correct tool name and parameters. Each task is executed with 5 independent trials to enable robust pass@k estimation.

The system prompt contains the format explanation, the tool definitions serialized in the target format, and a placeholder template showing the expected key layout of a tool call. The response is structured as plain-text reasoning followed by a \texttt{TOOL\_CALL} marker on its own line and then a single format-document containing the tool name and arguments. This layout keeps the format block at a known boundary so the model never has to switch formats mid-line, while still allowing free-form reasoning before the call. The text after \texttt{TOOL\_CALL} is parsed strictly in the target format and matched against ground truth.

The benchmark comprises 1,509 tasks spanning six domains, namely map, payment, file system, browser, search, and finance, exposing 87 distinct tools in total. Per-task tool counts range from a single function (finance) to 32 (map and browser), covering both low- and high-tool-count settings within a single benchmark.

Evaluation uses deterministic AST matching for tool-name correctness, with DeepSeek-V3 \citep{deepseek_v3_2024} serving as an LLM judge for parameter correctness where AST matching is insufficient.

\subsection{MCP-Universe}
MCP-Universe \citep{luo2025mcpuniverse} evaluates multi-turn agentic reasoning. While MCP-Universe ships with both a native function-calling mode (using the OpenAI tool-use API) and a text-based ReAct agent \citep{yao2023react}, we use the ReAct agent because the native API manages tool schemas internally and so cannot accept custom serializations. The ReAct agent embeds tool schemas as text in the prompt, giving us full control over the format. The agent iteratively reasons about the task, selects tools, executes them against live MCP servers, and incorporates results into its conversation history. Execution continues until the agent produces a final answer or reaches a maximum of 20 iterations.

The system prompt contains the format explanation, the tools serialized in the target format, and two worked response templates in that format: one showing a single ReAct step (thought, action, arguments) and one showing a final answer. The LLM is instructed to emit its entire per-turn response as one format-document, rather than wrapping the format-specific content inside labeled fields, so that strict parsing applies to the whole response.

Each task is evaluated through execution-based verification. Task-specific evaluator expressions check actual state changes (e.g., whether a financial calculation matches the expected value, whether a file was created with the correct content) rather than relying on output text comparison. This makes evaluation robust to format changes in the agent's responses, and no LLM judge is required.

The benchmark comprises 231 tasks across six domains that exercise distinct MCP server stacks, namely web search (SerpAPI), location navigation (Google Maps), financial analysis (yfinance, calculator), browser automation (Playwright, Chrome), repository management (GitHub, filesystem), and 3D design (Blender). This coverage ranges from flat API responses to deeply nested state and therefore stresses different aspects of format serialization. As noted in the Limitations, we excluded the web search category for this study, leaving 5 of 6 categories reported.

\subsection{BFCL}
The Berkeley Function Calling Leaderboard (BFCL v4), from the Gorilla project \citep{patil2023gorilla}, evaluates function calling with a deterministic Abstract Syntax Tree (AST) matcher that compares the predicted invocation against a reference. This avoids the cost and variance of an LLM judge and makes BFCL the most reproducible of the four benchmarks.

We integrated the shared serialization library into BFCL's prompt-mode pipeline by extending the handler and prompt builder to emit schemas in the target format and to parse tool calls in the target format. The system prompt contains the format explanation, the tools serialized in the target format, and a placeholder template showing the function-call array shape in that format. The LLM's entire response is the format-document and is parsed strictly. Separate model-registry entries per (model, format, scope) keep BFCL's per-model result cache from colliding across runs, and an environment-variable override (\texttt{BFCL\_PROMPT\_FORMAT\_OVERRIDE}) lets us switch formats without editing test data.

We restrict our evaluation to the \emph{non-live} and \emph{multi-turn} categories, which together cover the core single-call, multi-call, and multi-step scenarios with fully deterministic evaluation. The \emph{live} categories rely on real public APIs whose responses change over time, which we excluded to keep evaluation reproducible. The selected subset comprises roughly 2{,}040 tasks, combining single-function calls, parallel invocations, selection among multiple candidate functions, irrelevance-refusal tasks, and multi-turn scenarios with results, missing functions, missing parameters, and long contexts. Multi-turn categories only support input-only compression, as their evaluator hard-codes Python tool-call output.

\subsection{StableToolBench}
StableToolBench \citep{Guo_Cheng_Wang_Liang_Qin_Li_Liu_Sun_Liu_2025} is a reproducibility-focused variant of ToolBench \citep{qin2024toolllm} that evaluates multi-turn agentic reasoning against 7{,}000+ real-world APIs through a virtual API server. The virtual server returns cached responses for previously-seen API calls and falls back to a GPT-4-based simulator for cache misses, eliminating the reproducibility problems caused by drifting or retired upstream APIs in the original ToolBench. We use the pre-filtered solvable-queries subset, which retains only tasks confirmed to be answerable with the current virtual server.

We use a ReACT-style multi-turn agent \citep{yao2023react} and, as in MCP-Universe, do not use the native tool-call API because it cannot accept custom serializations. Earlier pilot runs used a textual \texttt{Thought:/Action:/Action Input:} shell with the format-specific content embedded inside the \texttt{Action Input:} field, but this collapsed adoption of TOON or TRON because the surrounding labels triggered the model's JSON-tool-call prior. We therefore restructured the prompt so that the LLM emits one full format-document per turn, with explicit \texttt{thought}, \texttt{action}, and \texttt{arguments} keys. The system prompt contains the format explanation, the tools serialized in the target format, and three worked response templates in that format covering a normal step, finishing the task, and giving up. The whole turn is parsed strictly as the target format.

StableToolBench's original evaluator (SoPR) relies on an LLM judge: we use GPT-4 as the judge for a subset of models and DeepSeek-R1-Distill-32B for the remainder, holding the judge fixed per model so that format comparisons are unaffected by judge choice.

The benchmark comprises 330 solvable queries across three difficulty groups of increasing scope. The easiest group targets single-tool queries with unseen instructions, the middle group combines multiple tools within the same category, and the hardest group requires coordinating tools across different categories.

\section{Models}
\label{app:models}

Table~\ref{tab:models} lists the five open-weight LLM configurations evaluated in this study, together with their parameter counts (total and active for the MoE configuration), whether the configuration emits an explicit chain-of-thought at inference time, the dominant post-training objective, and the architecture. Qwen3-32B is reported as two separate configurations with thinking on and off.

Generation parameters are benchmark-specific but kept identical across models within each benchmark for comparability: greedy decoding for BFCL, temperature 0.7 / top-p 1.0 for MCPToolBenchPP, temperature 0.6 with high-reasoning mode for MCP-Universe, and temperature 0.7 for StableToolBench (following the original ToolBench configuration).

\begin{table*}[h]
    \centering
    \footnotesize
    \caption{Open-weight LLM configurations evaluated. \textbf{Reasoning} indicates whether the configuration emits an explicit chain-of-thought at inference time. \textbf{Training emphasis} summarises the dominant post-training objective reported by each model's developers. Qwen3-32B is reported twice with thinking on and off as two separate configurations.}
    \label{tab:models}
    \begin{tabular}{@{}lrrcll@{}}
        \toprule
        \textbf{Model} & \textbf{Total} & \textbf{Active} & \textbf{Reasoning} & \textbf{Training emphasis} & \textbf{Architecture} \\
        \midrule
        Mistral-Small-24B-Instruct      & 24B  & 24B & $\times$        & tool calling & dense, AWQ \\
        Qwen3-32B (thinking on)         & 32B  & 32B & \checkmark      & tool calling & dense, AWQ \\
        Qwen3-32B (thinking off)        & 32B  & 32B & $\times$        & tool calling & dense, AWQ \\
        DeepSeek-R1-Distill-Qwen-32B    & 32B  & 32B & \checkmark      & reasoning    & dense, R1-distilled, AWQ \\
        Llama-4-Scout-17B-16E           & 109B & 17B & $\times$        & general      & MoE (16 experts), w4a16 \\
        \bottomrule
    \end{tabular}
\end{table*}

\section{Implementation Notes}
\label{app:implementation_notes}

\paragraph{TRON port to Python.} The TRON reference implementation \citep{tron-format/tron-javascript_2026} is provided in JavaScript. Since our benchmark pipelines are written in Python, we ported the implementation to Python while preserving the original semantics, and wrap it behind the shared\_format library together with the official TOON Python library.

\paragraph{Model-specific output-parsing fixes.} Pilot runs surfaced two model-specific parsing issues that, if uncorrected, silently zero out entire (model, format) sweeps in the strict-parsing pipeline. We resolved each at the parser layer before the reported runs.

\begin{itemize}
    \item \textbf{Reasoning-prefix passthrough (Qwen3 thinking-on, DeepSeek-R1-Distill).} Both models emit a leading \texttt{<think>...</think>} block that the strict ReAct parser treats as malformed prefix content. We strip the block before parsing in \texttt{react.py} and in the BFCL handler utilities.
    \item \textbf{Markdown code-fence wrapping (Llama-4-Scout).} Llama-4-Scout wraps its response in \texttt{```json ... ```} fences even when prompted for plain target-format output. The original parser anchored at the start of the response and failed on the fence. We now extract the first fenced block and parse its contents.
\end{itemize}

These fixes are uniform across formats and benchmarks, so they do not bias any (format, format) comparison. They illustrate that strict-parsing pipelines in agentic evaluations have a non-trivial model-specific surface that benchmarks should publish alongside their headline numbers.

\section{Serialization Examples}
\label{app:serialization_examples}

Figure~\ref{fig:serialization_examples} illustrates the structural differences between the three formats on identical sample data. We evaluate three serialization formats.

\begin{figure*}[t]
\centering
\scriptsize

\begin{minipage}[t]{0.32\textwidth}
\textbf{JSON}
\vspace{2mm}

\begin{verbatim}
{
  "context": {
    "task": "Our favorite hikes",
    "location": "Boulder",
    "season": "spring_2025"
  },
  "friends": ["ana", "luis", "sam"],
  "hikes": [
    {
     "id": 1, 
     "name": "Blue Lake Trail",
     "distanceKm": 7.5
    },
    {
     "id": 2, 
     "name": "Ridge Overlook",
     "distanceKm": 9.2
    },
    {
     "id": 3, 
     "name": "Wildflower Loop",
     "distanceKm": 5.1
    }
  ]
}
\end{verbatim}
\end{minipage}
\hfill
\begin{minipage}[t]{0.32\textwidth}
\textbf{TOON}
\vspace{2mm}

\begin{verbatim}
context:
  task: Our favorite hikes
  location: Boulder
  season: spring_2025

friends[3]: ana,luis,sam

hikes[3]{id,name,distanceKm}:
  1,Blue Lake Trail,7.5
  2,Ridge Overlook,9.2
  3,Wildflower Loop,5.1
\end{verbatim}
\end{minipage}
\hfill
\begin{minipage}[t]{0.32\textwidth}
\textbf{TRON}
\vspace{2mm}

\begin{verbatim}
class A: id,name,distanceKm

{"context":{"task":"Our favorite
hikes","location":"Boulder","season":
"spring_2025"},"friends":["ana","luis",
"sam"],"hikes":[A(1,"Blue Lake Trail",
7.5),A(2,"Ridge Overlook",9.2),A(3,
"Wildflower Loop",5.1)]}
\end{verbatim}
\end{minipage}

\caption{Example encodings of the same data in JSON, TOON, and TRON.}
\label{fig:serialization_examples}

\end{figure*}

\section{Per-benchmark Facets across All Models}
\label{app:faceted_all}

\begin{figure*}[!htbp]
    \centering
    \includegraphics[width=0.95\textwidth]{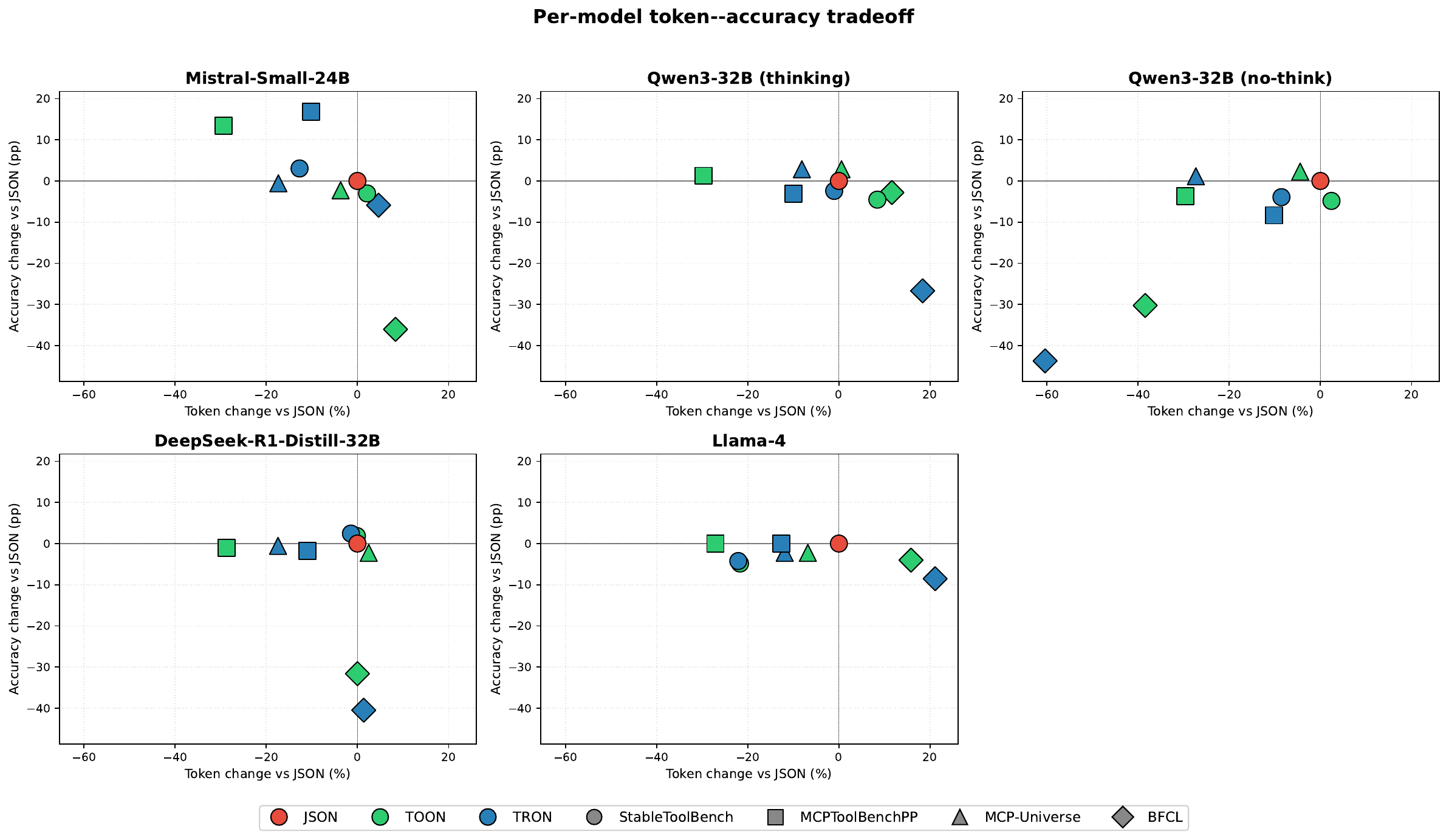}
        \caption{Per-benchmark token--accuracy facets under input-only compression for all five model configurations. Each panel reports one model, with markers per (format, benchmark); JSON sits at the origin by construction. The dominant pattern is per-benchmark, not per-format: BFCL drops substantially on accuracy for most (model, format) combinations, with reasoning configurations (Qwen3-32B both modes, DeepSeek-R1) showing TRON drops of $-27$ to $-44$~pp. MCPToolBenchPP clusters on the left ($\approx -30\%$ tokens under TOON, $\approx -10\%$ under TRON) with mixed accuracy effects, including Mistral-Small-24B's positive gain. Multi-turn benchmarks sit nearer the origin on both axes, with TOON exhibiting the parsing-cascade effect on MCP-Universe and StableToolBench.}
    \label{fig:faceted_scatter_all}
\end{figure*}

Figure~\ref{fig:faceted_scatter_all} reports the per-benchmark token--accuracy facets under input-only compression for all five model configurations, broken down by benchmark. The all-models view exposes the per-benchmark variation that the cross-benchmark means in Section~\ref{sec:results} aggregate over, including the BFCL accuracy collapse for smaller and non-reasoning models, the counter-intuitive accuracy gain on MCPToolBenchPP for Mistral-Small-24B, and the relative stability of TRON on multi-turn benchmarks.

\section{Per-cell Accuracy and Token Change}
\label{app:per_cell}

Tables~\ref{tab:accuracy_input_only} and~\ref{tab:accuracy_full} report the per-(benchmark, model, format) accuracy and total-token change relative to the JSON baseline that back the aggregate claims in Sections~\ref{sec:results} and~\ref{sec:discussion}. The two tables differ only in the experimental condition (input-only compression vs full compression). Values are sourced from \texttt{results\_final/data\_real.json}.

\begin{table*}[!htbp]
\centering
\footnotesize
\setlength{\tabcolsep}{4pt}
\caption{Per-cell accuracy and token change under \textbf{input-only} compression. JSON cells report accuracy only (token change is zero by definition). TOON and TRON cells report accuracy~/~$\Delta$tokens vs JSON, both in percent. Llama-4-Scout returns 0.0 on MCPToolBenchPP across all formats and on MCP-Universe under TOON and TRON. These are real measurements reflecting the model's failure to follow the prompt protocol on those stacks, not missing data.}
\label{tab:accuracy_input_only}
\begin{tabular}{@{}llrrrrr@{}}
\toprule
\textbf{Benchmark} & \textbf{Fmt} & \textbf{Mistral} & \textbf{Qwen3(t)} & \textbf{Qwen3(n)} & \textbf{DeepSeek-R1} & \textbf{Llama-4} \\
\midrule
\multirow{3}{*}{BFCL} & JSON & 88.8 & 93.0 & 93.3 & 82.7 & 85.5 \\
 & TOON & 52.7~/~$+8.4$ & 90.2~/~$+11.6$ & 63.1~/~$-38.4$ & 51.1~/~$-0.0$ & 81.5~/~$+15.8$ \\
 & TRON & 82.9~/~$+4.6$ & 66.3~/~$+18.4$ & 49.6~/~$-60.4$ & 42.3~/~$+1.4$ & 76.9~/~$+21.1$ \\
\midrule
\multirow{3}{*}{MCPToolBenchPP} & JSON & 8.3 & 30.7 & 30.8 & 29.5 & 0.0 \\
 & TOON & 21.8~/~$-29.3$ & 31.9~/~$-29.7$ & 27.2~/~$-29.7$ & 28.5~/~$-28.7$ & 0.0~/~$-27.1$ \\
 & TRON & 25.1~/~$-10.2$ & 27.5~/~$-10.0$ & 22.5~/~$-10.2$ & 27.7~/~$-10.9$ & 0.0~/~$-12.7$ \\
\midrule
\multirow{3}{*}{MCP-Universe} & JSON & 4.0 & 5.7 & 6.8 & 9.7 & 2.3 \\
 & TOON & 1.7~/~$-3.7$ & 8.5~/~$+0.6$ & 9.1~/~$-4.4$ & 7.4~/~$+2.5$ & 0.0~/~$-6.8$ \\
 & TRON & 3.4~/~$-17.3$ & 8.5~/~$-8.1$ & 8.0~/~$-27.3$ & 9.1~/~$-17.4$ & 0.0~/~$-11.9$ \\
\midrule
\multirow{3}{*}{StableToolBench} & JSON & 20.0 & 32.4 & 33.6 & 20.0 & 29.7 \\
 & TOON & 17.0~/~$+2.1$ & 27.9~/~$+8.4$ & 28.8~/~$+2.4$ & 21.8~/~$-0.1$ & 24.9~/~$-21.7$ \\
 & TRON & 23.0~/~$-12.7$ & 30.0~/~$-1.0$ & 29.7~/~$-8.5$ & 22.4~/~$-1.4$ & 25.4~/~$-22.1$ \\
\bottomrule
\end{tabular}
\end{table*}

\begin{table*}[!htbp]
\centering
\footnotesize
\setlength{\tabcolsep}{4pt}
\caption{Per-cell accuracy and token change under \textbf{full} compression. JSON cells report accuracy only (token change is zero by definition). TOON and TRON cells report accuracy~/~$\Delta$tokens vs JSON, both in percent. Llama-4-Scout returns 0.0 on MCPToolBenchPP across all formats and on MCP-Universe under TOON and TRON. These are real measurements reflecting the model's failure to follow the prompt protocol on those stacks, not missing data.}
\label{tab:accuracy_full}
\begin{tabular}{@{}llrrrrr@{}}
\toprule
\textbf{Benchmark} & \textbf{Fmt} & \textbf{Mistral} & \textbf{Qwen3(t)} & \textbf{Qwen3(n)} & \textbf{DeepSeek-R1} & \textbf{Llama-4} \\
\midrule
\multirow{3}{*}{BFCL} & JSON & 88.8 & 93.0 & 93.3 & 82.7 & 85.5 \\
 & TOON & 57.1~/~$+9.8$ & 57.7~/~$+14.6$ & 57.7~/~$+14.1$ & 56.5~/~$+0.7$ & 55.9~/~$+21.2$ \\
 & TRON & 64.5~/~$+8.3$ & 52.2~/~$+21.6$ & 28.6~/~$-55.3$ & 41.0~/~$-20.1$ & 55.0~/~$+24.6$ \\
\midrule
\multirow{3}{*}{MCPToolBenchPP} & JSON & 8.3 & 30.7 & 30.8 & 29.5 & 0.0 \\
 & TOON & 13.8~/~$-30.6$ & 16.4~/~$-30.6$ & 20.3~/~$-31.0$ & 15.8~/~$-30.2$ & 0.0~/~$-28.8$ \\
 & TRON & 25.3~/~$-11.6$ & 25.8~/~$-10.9$ & 23.4~/~$-11.4$ & 27.8~/~$-11.7$ & 0.0~/~$-14.4$ \\
\midrule
\multirow{3}{*}{MCP-Universe} & JSON & 4.0 & 5.7 & 6.8 & 9.7 & 2.3 \\
 & TOON & 2.3~/~$+1.7$ & 1.1~/~$+41.0$ & 1.7~/~$+41.2$ & 1.7~/~$+13.1$ & 0.6~/~$-20.7$ \\
 & TRON & 2.3~/~$-8.2$ & 8.0~/~$-16.3$ & 7.4~/~$-18.8$ & 5.1~/~$+18.2$ & 0.0~/~$-13.6$ \\
\midrule
\multirow{3}{*}{StableToolBench} & JSON & 20.0 & 32.4 & 33.6 & 20.0 & 29.7 \\
 & TOON & 11.5~/~$+14.4$ & 24.2~/~$+7.7$ & 19.4~/~$+21.0$ & 46.4~/~$-14.9$ & 14.8~/~$-3.3$ \\
 & TRON & 27.6~/~$-9.3$ & 19.1~/~$+49.2$ & 23.0~/~$+21.7$ & 20.9~/~$-17.6$ & 24.6~/~$-27.9$ \\
\bottomrule
\end{tabular}
\end{table*}

\section{Absolute-Sum Token--Accuracy View}
\label{app:absolute_sum}

\begin{figure*}[!htbp]
    \centering
    \includegraphics[width=0.95\textwidth]{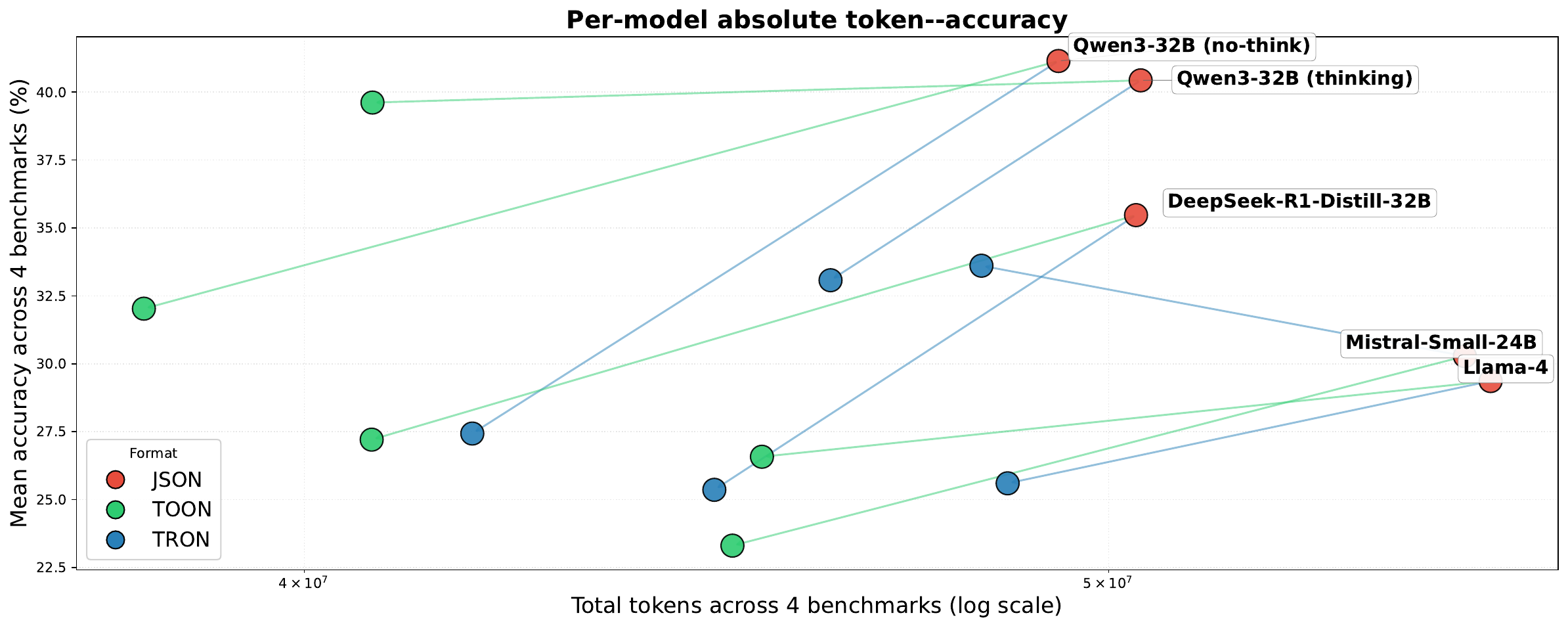}
    \caption{Per-model absolute token--accuracy view, summed across the four benchmarks. Each marker is one (model, format) combination, with the x axis showing total tokens on a log scale and the y axis showing mean accuracy across benchmarks. Arrows trace the JSON $\to$ TOON and JSON $\to$ TRON movement per model. Under this absolute-sum aggregation, TOON arrows reach further along the token axis (18--22\% reduction per model) than TRON arrows (8--15\%), because MCPToolBenchPP (which TOON compresses by roughly 30\%) dominates the total token budget. The per-cell numbers behind this view are tabulated in Table~\ref{tab:accuracy_input_only}.}
    \label{fig:absolute_scatter_wide}
\end{figure*}

Figure~\ref{fig:pipeline_overview}c in the main body uses a mean-of-percentages aggregation on the x-axis (each benchmark contributes equally to the per-model summary). Figure~\ref{fig:absolute_scatter_wide} reports the same per-model view under an absolute-sum aggregation, in which the x-axis is total tokens summed across the four benchmarks on a log scale. The two views can disagree on which format is more compressive for a given model when the benchmarks have very different token volumes (MCPToolBenchPP alone accounts for roughly 70\% of total tokens in our matrix), so a format that wins big on a low-volume benchmark in the per-percentage view may lose to a format that wins moderately on a high-volume benchmark in the absolute-sum view. We retain the mean-of-percentages aggregation as the headline view because it weights each benchmark equally and matches standard NLP-evaluation practice; the absolute-sum view is reported here for transparency.

\end{document}